\documentclass[10pt,twocolumn,letterpaper]{article}

\usepackage[utf8]{inputenc}
\usepackage{iccv}
\usepackage{times}
\usepackage{epsfig}
\usepackage{graphicx}
\usepackage{amsmath}
\usepackage{amssymb}
\usepackage{caption}
\graphicspath{ {figs/} }
\usepackage{multirow}
\usepackage{subcaption}

\usepackage{soul} 

\usepackage{verbatim} 

\usepackage[pagebackref=true,breaklinks=true,letterpaper=true,colorlinks,bookmarks=false]{hyperref}

\iccvfinalcopy 

\def\iccvPaperID{2769} 

\ificcvfinal\fi

\author{Yubo Zhang\\
Carnegie Mellon University\\
Pittsburgh, PA 15213, USA\\
{\tt\small yuboz@andrew.cmu.edu}
\and
Vishnu Naresh Boddeti\\
Michigan State University\\
East Lansing, MI 48824, USA\\
{\tt\small vishnu@msu.edu}
\and
Kris M. Kitani\\
Carnegie Mellon University\\
Pittsburgh, PA 15213, USA\\
{\tt\small kkitani@cmu.edu}
}

\begin{document}

\title{Gesture-based Bootstrapping for Egocentric Hand Segmentation}

\maketitle

\begin{abstract}
Accurately identifying hands in images is a key sub-task for human activity understanding with wearable first-person point-of-view cameras. Traditional hand segmentation approaches rely on a large corpus of manually labeled data to generate robust hand detectors. However, these approaches still face challenges as the appearance of the hand varies greatly across users, tasks, environments or illumination conditions. A key observation in the case of many wearable applications and interfaces is that, it is only necessary to accurately detect the user's hands in a specific situational context. Based on this observation, we introduce an interactive approach to learn a person-specific hand segmentation model that does not require any manually labeled training data. The training in our approach proceeds in two steps, an interactive bootstrapping step for identifying moving hand regions, followed by learning a personalized user specific hand appearance model. Concretely, our approach uses two convolutional neural networks at training stage: (1) a trained gesture network that uses pre-defined motion information to detect the hand regions which are delivered to the next network for training the hand detector; and (2) an appearance network that learns a person specific model of the hand region based on the output of the gesture network. During training, to make the appearance network robust to errors in the gesture network, the loss function of the former network incorporates the confidence of the gesture network while learning. Experiments demonstrate the robustness of our approach with an $F_1$ score over 0.8 on all challenging datasets across a wide range of illumination and hand appearance variations, improving upon a baseline approach by more than 10\%.
\end{abstract}


\section{Introduction}
We address the task of pixel-level hand detection using first person wearable cameras. From the perspective of human activity understanding, pixel-level detection of hand and arm regions is an important sub-task for higher levels of human activity understanding using a wearable camera. The exact shape and transformation of the hand holds important information about the task that is being performed. In the context of virtual or augmented reality systems, the precise detection of hand regions is needed for realistic rendering and interactions with virtual objects. In the context of patient monitoring for neuromuscular rehabilitation, the exact hand shape (\eg grasp or manipulation) is needed to accurately characterize the wearer's motor skills.

\begin{figure}[t]
    \centering
    \includegraphics[width=1.0\linewidth]{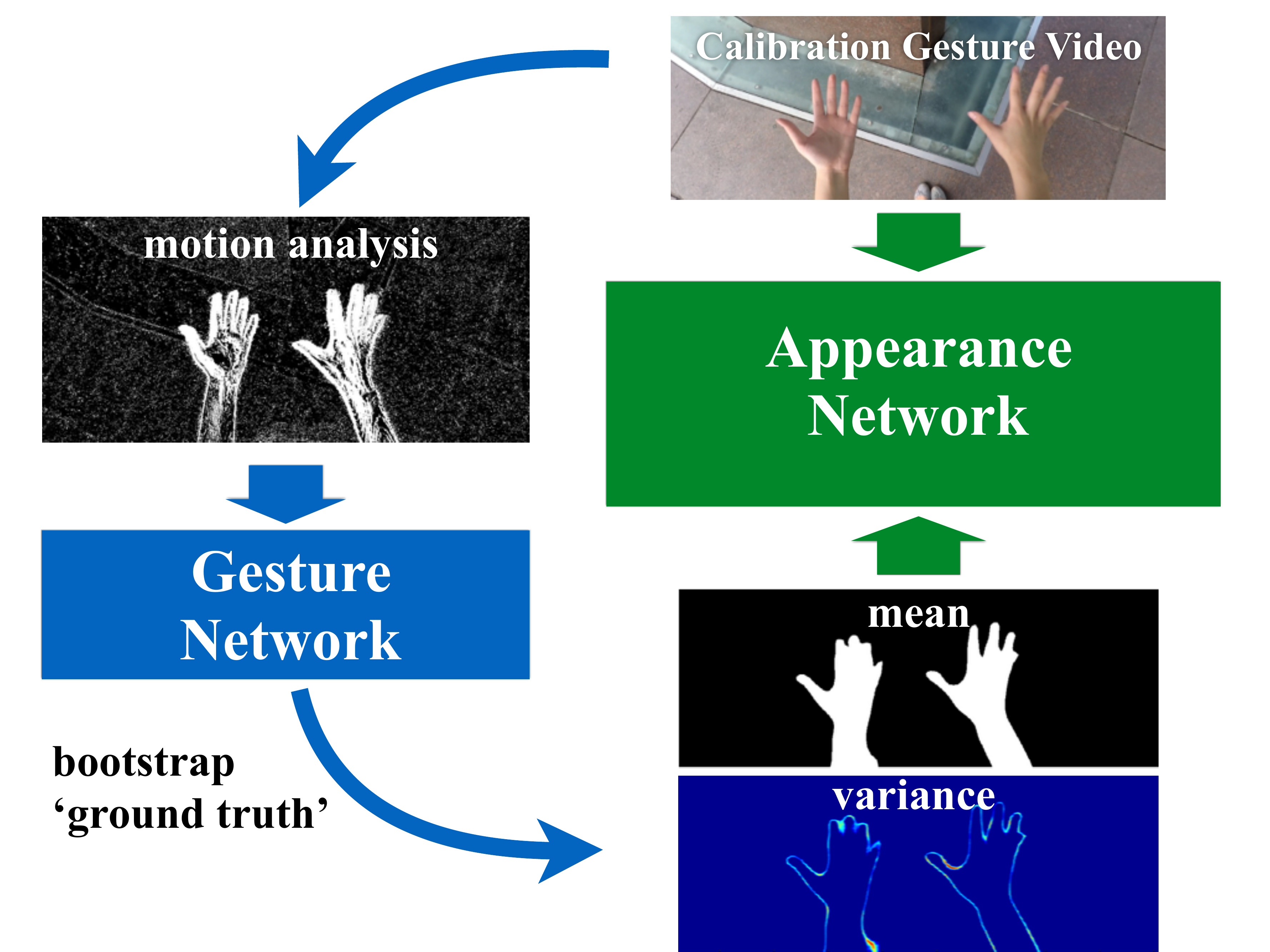}
    \caption{Our method bootstraps ground truth labels using motion analysis to train a personalized hand segmentation network.}
    \label{fig:teaser}
\end{figure}

\begin{figure*}[t]
\center
\includegraphics[width=28mm,height=15mm]{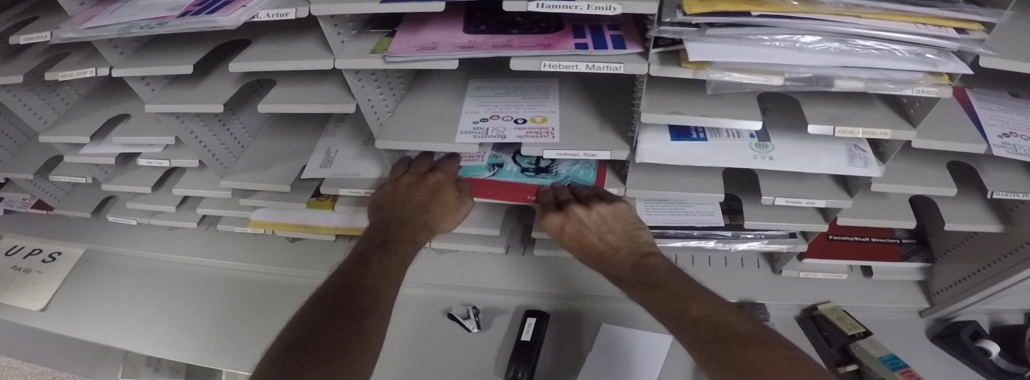}
\includegraphics[width=28mm,height=15mm]{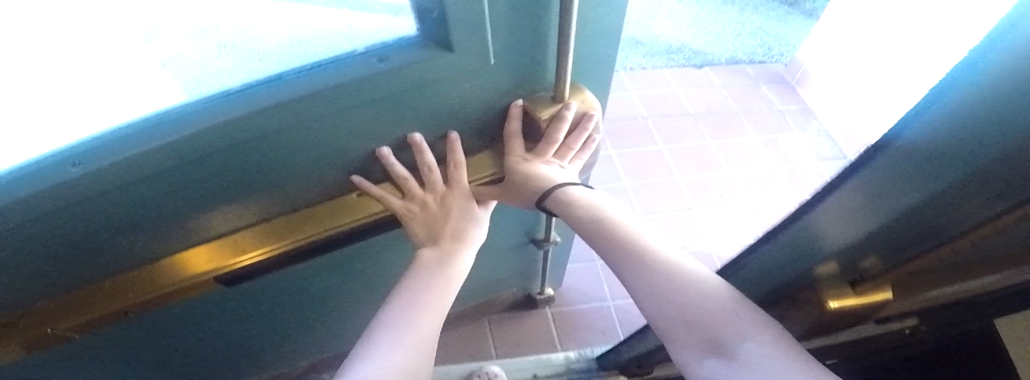}
\includegraphics[width=28mm,height=15mm]{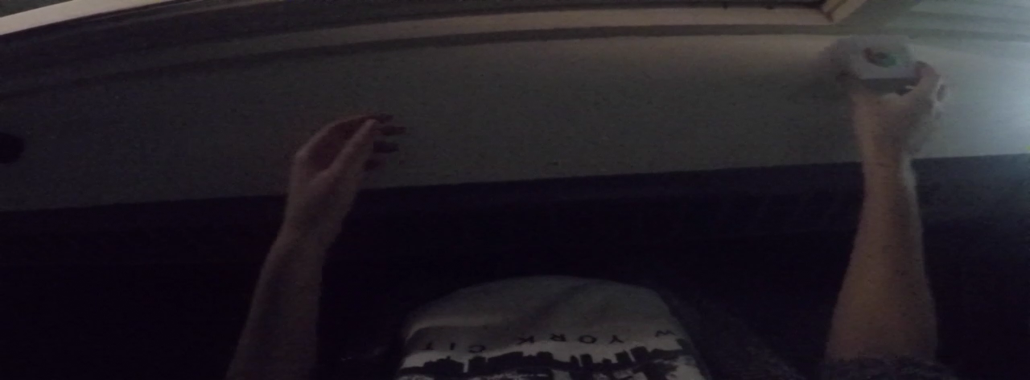}
\includegraphics[width=28mm,height=15mm]{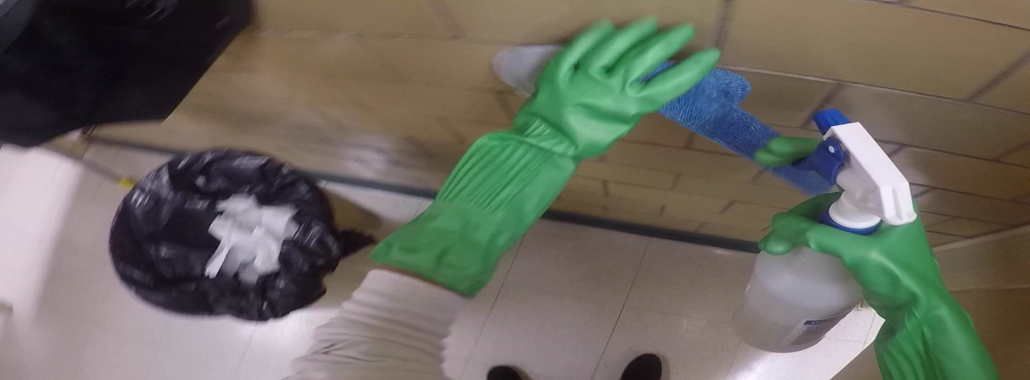}
\includegraphics[width=28mm,height=15mm]{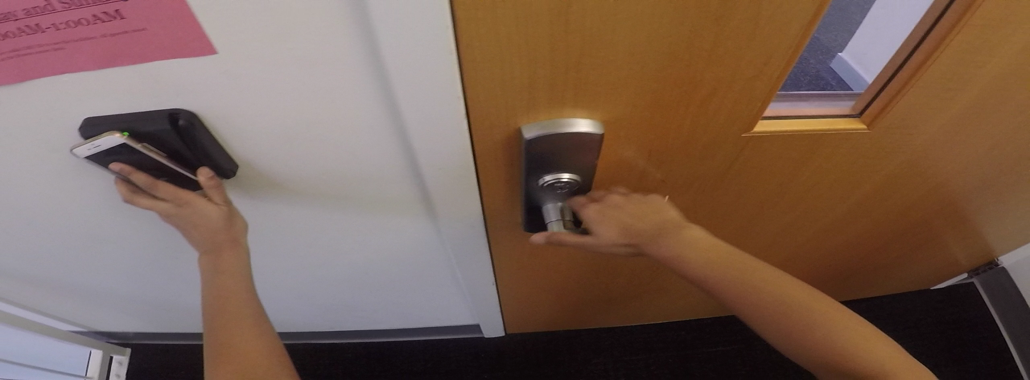}
\includegraphics[width=28mm,height=15mm]{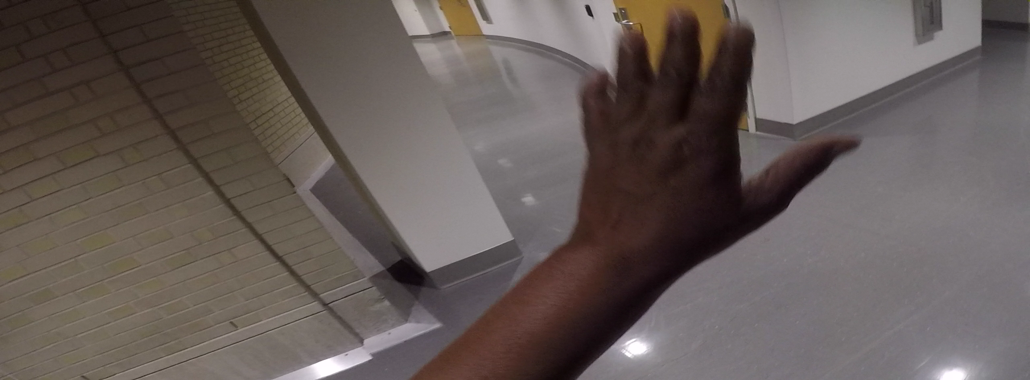}
\\
\includegraphics[width=28mm,height=15mm]{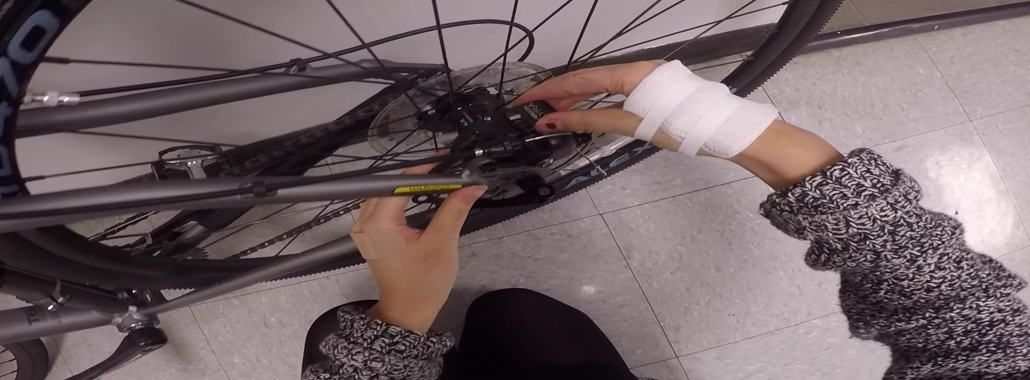}
\includegraphics[width=28mm,height=15mm]{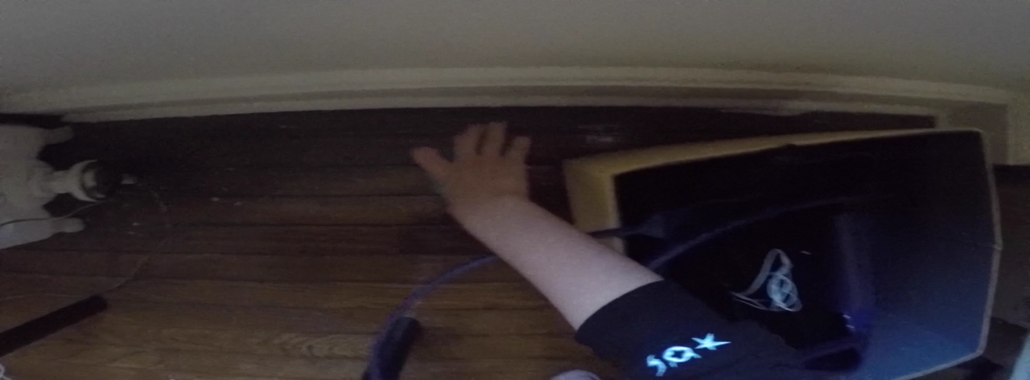}
\includegraphics[width=28mm,height=15mm]{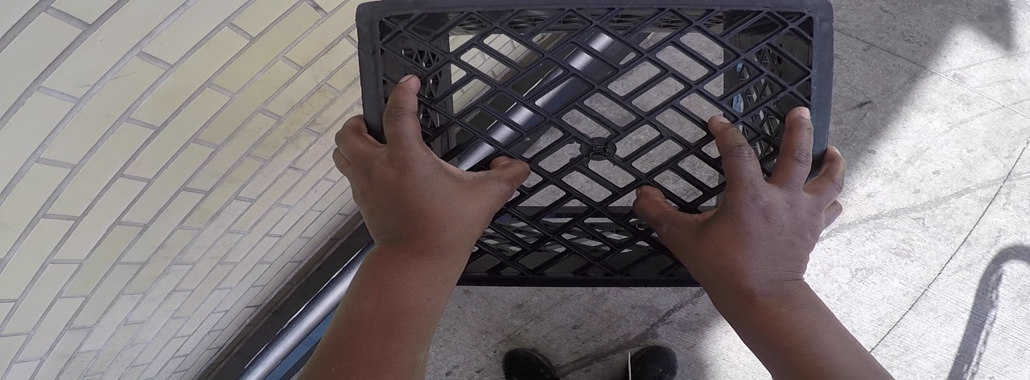}
\includegraphics[width=28mm,height=15mm]{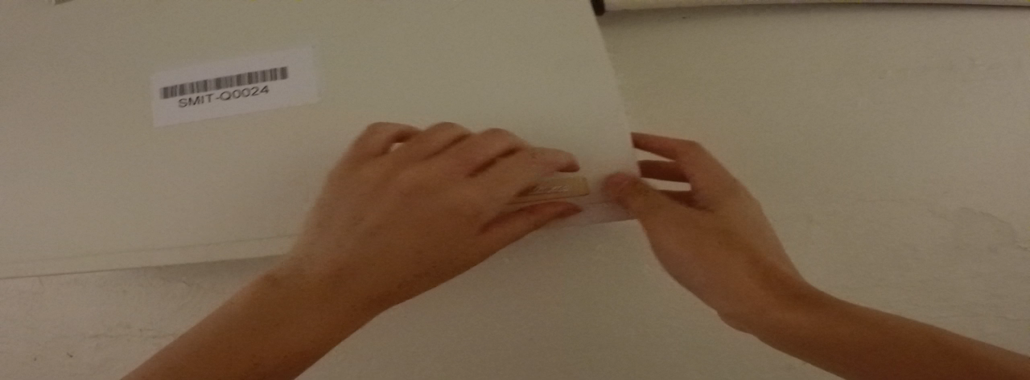}
\includegraphics[width=28mm,height=15mm]{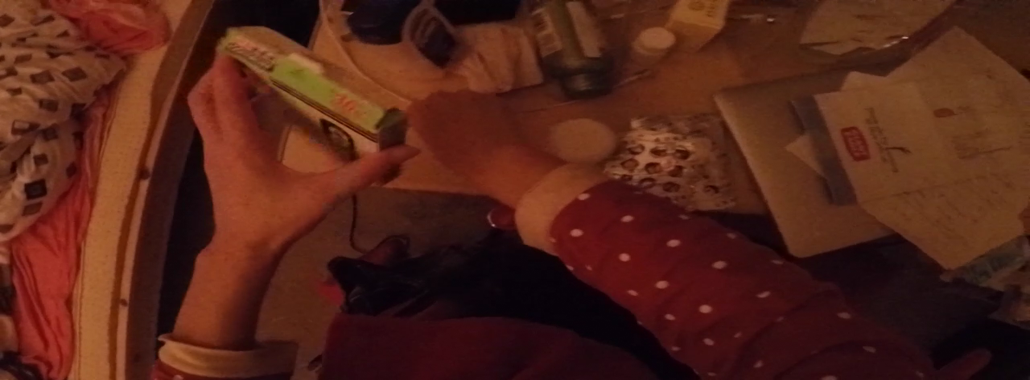}
\includegraphics[width=28mm,height=15mm]{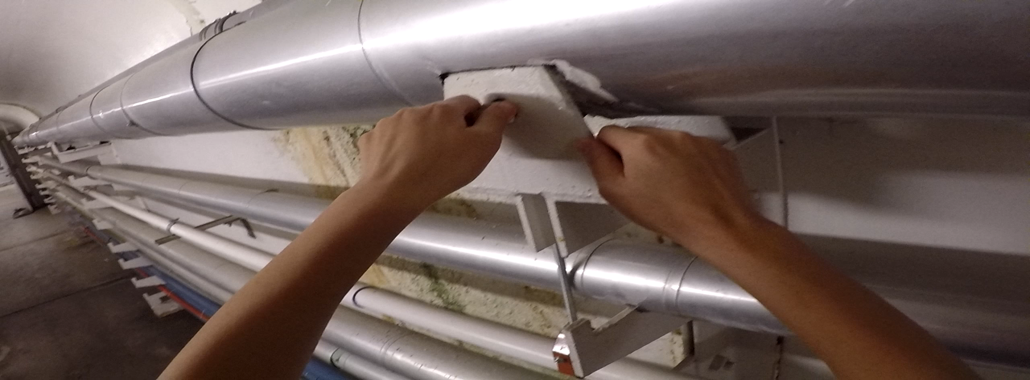}
\caption{Sample images from our proposed dataset egocentric hand segmentation dataset. The dataset spans a diverse set of users across a variety of environments spanning multiple backgrounds and illumination variations.}
\label{fig:dataset}
\end{figure*}

The task of building a model to detect hand regions with wearable cameras is daunting. First, the images of hands captured by first-person point-of-view cameras can exhibit very large variations in appearance. In general,there are no constraints on the environment in which the wearable camera might be used. The environment might include videos of a person playing a game in a dark room or playing outdoor sports under the bright sun. Training a single hand detector that can detect hands under all possible illumination conditions may be possible but would require a tremendous amount of labeled data. Second, the collection of pixel-level labels is labor intensive. Even with a large corpus of hand images from every possible scenario, labeling such data is currently not feasible. A single image of the hands at standard resolution can contain close to 1 million pixels that must be manually labeled and can take several minutes to label. Based on these two challenges, it seems that training a hand detector that works for any person in any scenario will be quite challenging.

But do we really need to train a hand detector that works for all people and under any scenario? Is there anything else we can do to bootstrap a hand detection without gathering so much data? When we consider many of the use cases of wearable cameras, we make the two observations that hold the key to a more scalable approach. First, we observe that in many cases the hand detector usually only needs to work for one person -- the camera wearer. In some cases, it only needs to work in a limited environment, such as the home for rehabilitation or a single room for a VR application. In other words, it is completely acceptable to learn a personalized hand detector. Second, we make the observation that in applications such as VR and AR, the sensing of the hands is part of an interactive process, which also means that it is plausible to assume a certain level of user interaction and cooperation to help generate a better hand detection algorithm.

We take advantage of the appropriateness of personalization and the possibility of interaction, and propose a two stage deep network for hand detection (see Figure \ref{fig:teaser}). The detector will be optimized for the user and will not require any manual image labeling. The training process is as follows. (1) The user with the wearable camera is instructed to perform a simple hand gesture. (2) Using optical flow and background subtraction results as input, a pre-trained deep 'gesture' network model is used to estimate a foreground map (pixel-level mask of the hand and arm). (3) Images of the hand during the gesture interaction phase along with foreground maps are used to train a person-specific deep 'appearance' network. To deal with possible errors in the output of the gesture network, the loss function of the appearance network is weighted by the confidence of the gesture network.

\noindent\textbf{Contributions.} (1) We propose human interaction as a means of bootstrapping a hand segmentation algorithm for wearable cameras. (2) We show how the uncertainty of the foreground masks generated by the gesture network can be incorporated to adaptively weight the loss function of the appearance network. (3) We present a novel dataset (Figure \ref{fig:dataset}) with 4200 images of 10 users across 30 environments, along with the respective calibration gesture, as a benchmark dataset for egocentric hand segmentation.


\section{Previous work}

In prior work \cite{Fathi:2011:UEA:2355573.2356302} \cite{conf/eccv/FathiLR12} \cite{ren2010figure}, hand motion was considered as an important clue while tackling the hand segmentation problem in egocentric RGB videos. Several foreground-background segmentation approaches have been developed with the assumption that the background is static. Ren \etal \cite{ren2010figure} addressed the hand, and handled object segmentation tasks by solving figure-ground segmentation problem. They computed dense optical flow and separated the background with consistent motion patterns. Fathi \etal \cite{fathi2011learning}further solved the problem by estimating background model in short video intervals. While motion based methods are naturally robust to the appearance variance and do not require training in advance, they are inevitably error-prone in the situation where no hand motion is made or large body motion causes background to change significantly. Furthermore, it is hard to separate hands from the handled objects to get the region only with hands.

Unlike motion-based methods, appearance-based methods \cite{conf/cvpr/LeeGG12} \cite{DBLP:conf/iccv/LiFR13} \cite{ma2016going} \cite{Zariffa2013}are not constrained by extreme ego-motion of first-person cameras as these methods train hand detectors based on appearance features. Li and Kitani \cite{li2013pixel}were among the first to investigate different local appearance features for hand segmentation in first person vision. Serra \etal \cite{serra2013hand} investigated features that integrate illumination, temporary and spatial consistency, and further improved the performance of hand segmentation. Xiaolong \etal \cite{10.1109/WACV.2016.7477665} further proposed a two-stage hand detection algorithm which draws bounding boxes around hands and trains discriminative classification model with convolutional neural network. The appearance-based methods are robust to noisy ego-motion and motion blur since they only rely on the appearance of hands. However, these approaches do not generalize well to significant differences in skin colors and abrupt illumination changes. To handle complex changes in hand appearance, Li and Kitani \cite{li2013model} proposed recommendation system paradigm to select proper models for particular tasks; However, the performance is limited since it is impossible to cover all possible tasks in their probe.

As discussed above, much of the work in hand detection from first person vision has focused on extending traditional methods, with little focus on the potential to leverage human interaction as part of the bigger picture of an interactive wearable system. As one of the first examples introducing gesture based calibration, \cite{kumar2015fly} showed that a robust color space mapping combined with optical flow and region growing can be utilized to generate hand region detection which provides information to train a color based hand detector. Although the detection of hands in gesture calibration step using appearance and motion is unsupervised, it still makes prior assumptions about the color of the hands as a pre-processing step \cite{vandeWeijer:2009:LCN:1657314.1657325} and cannot handle large variations away from standard skin color. Furthermore, the appearance-based detection algorithm uses a parametric Gaussian mixture model whose parameters in practice require manual tuning when the environment changes.

With development of portable depth sensors, some work that has been done with RGBD images shows promising results in the field of egocentric object recognition \cite{Lin2015} \cite{conf/cvpr/WanA15} and activity recognition \cite{moghimi2014experiments} \cite{rogez2015understanding}. Rogez \etal \cite{rogez20143d} \cite{rogez2015first} focused on the task of 3D hand pose detection and trained the detector with depth information using synthetic training exemplars. They also showed that depth cues are particularly informative in the near-field first person viewpoints. \cite{moghimi2014experiments} \cite{conf/cvpr/WanA15} further showed that foreground segmentation is efficient enough to separate the hand and handled object area by combining RGB image with depth data. Lin \etal \cite{Lin2015} built a generic skin-tone model with histogram of oriented normal vectors (HONV) features that works well with 3D point cloud. Useful as the depth information is, current portable RGB-D sensors, like stereo cameras and cameras with structured light, have drawbacks on large power consumption. The NIR structured light also suffers from low accuracy in outdoor environment.

In this paper, we develop our method with RGB images for applicability. We leverage the benefit of human interaction by introducing a supervised motion-based hand segmentation network to obtain information from the particular user. The network only needs to be trained once and is very robust to appearance variations caused by skin tones and illumination changes. The information is then fed into the appearance-based detection algorithm which is trained discriminatively, robust to unconstrained hand movement, and does not require manual tuning of parameters across different environments or users.

\begin{figure}[tb]
    \centering
    \includegraphics[width=1.7\linewidth]{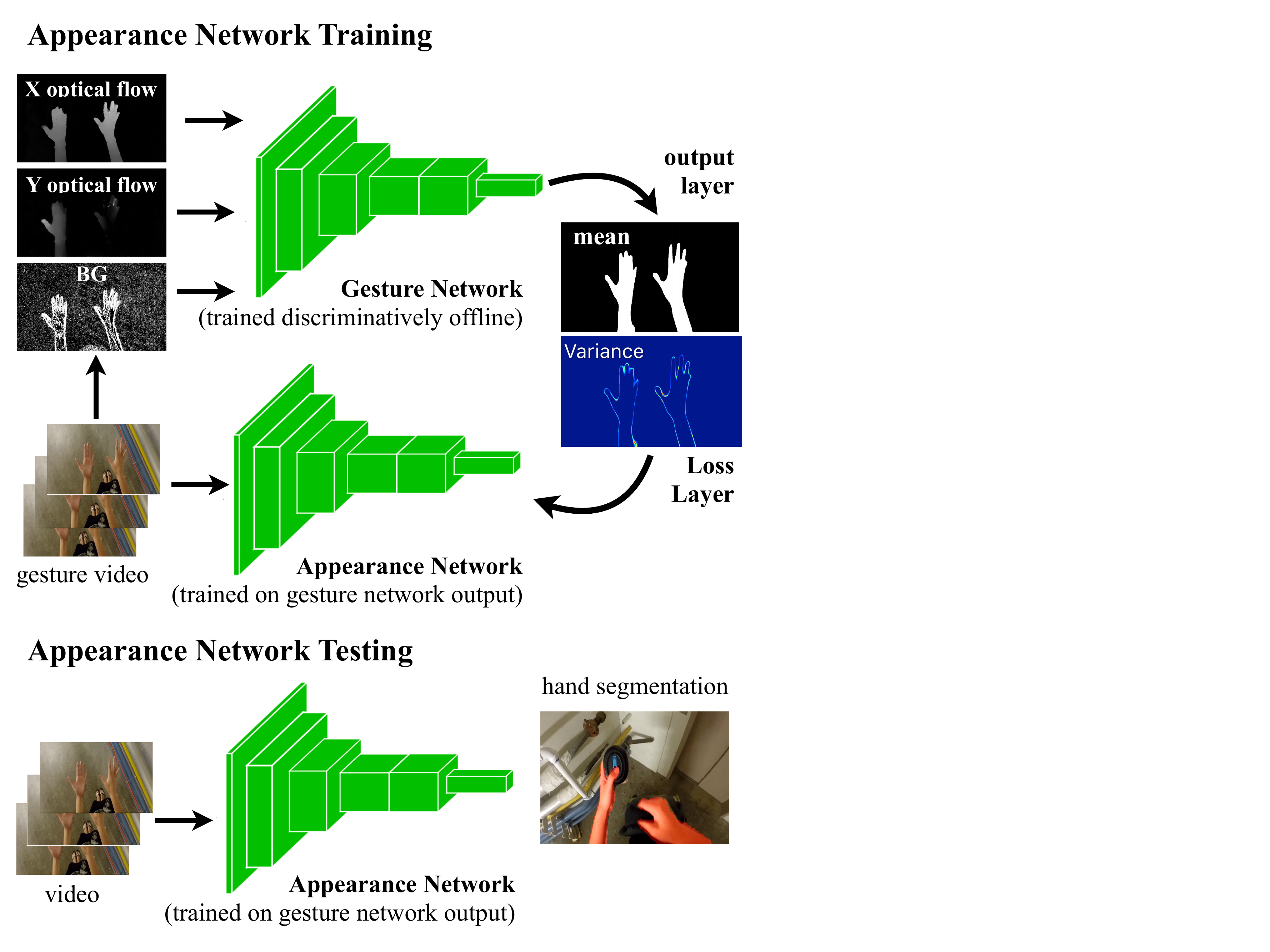}
    \caption{\textbf{Overview:} Our proposed two stage process for interactively learning personalized hand detection models. The person agnostic gesture network is designed for hand segmentation by analyzing the motion features of the interactive hand gesture. A person specific appearance network is then learned by bootstrapping the output of the gesture network as a proxy of the ground truth segmentation masks. The appearance network is designed to be robust to the errors of the gesture network by incorporating the confidence of the gesture network into its loss function. At test time the appearance network can then be used for hand segmentation across any unconstrained hand object interactions.}
    \label{fig:network}
\end{figure}


\begin{figure*}[!h]
    \centering
\includegraphics[width=30mm,height=15mm]{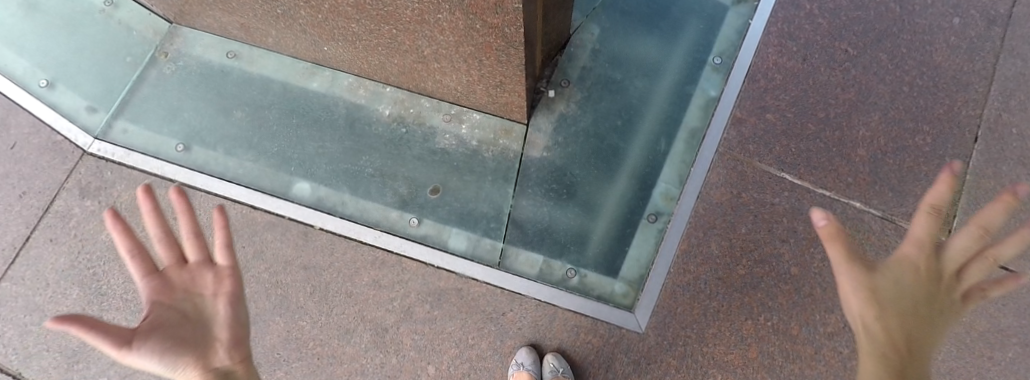}
\includegraphics[width=30mm,height=15mm]{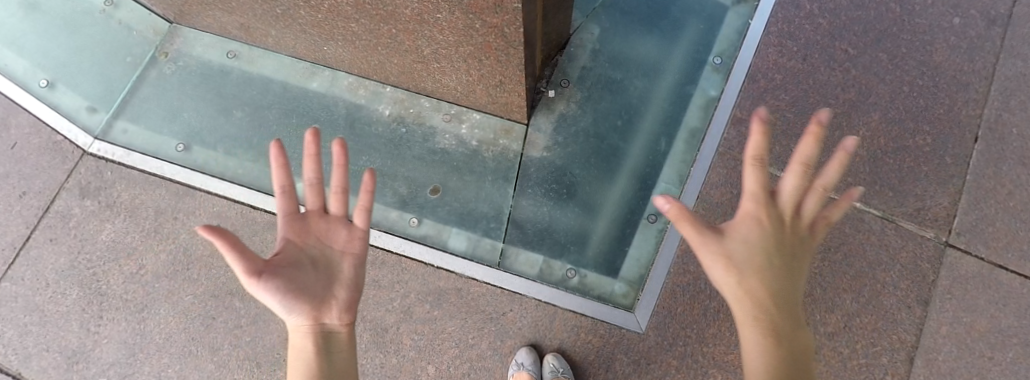}
\includegraphics[width=30mm,height=15mm]{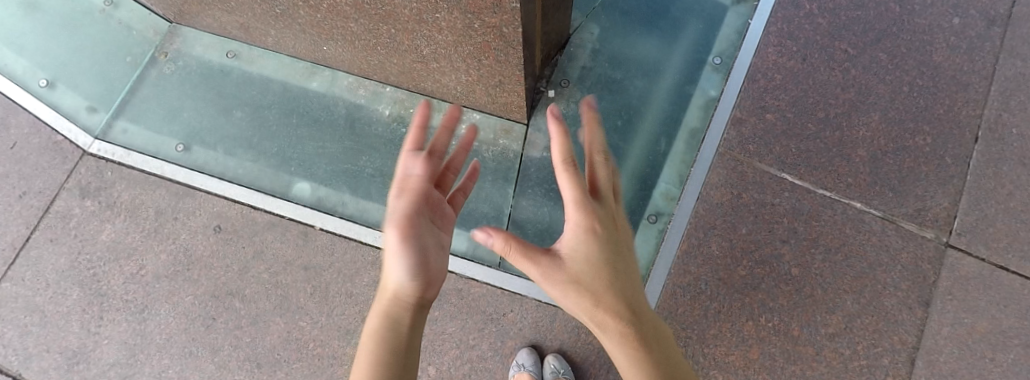}
\includegraphics[width=30mm,height=15mm]{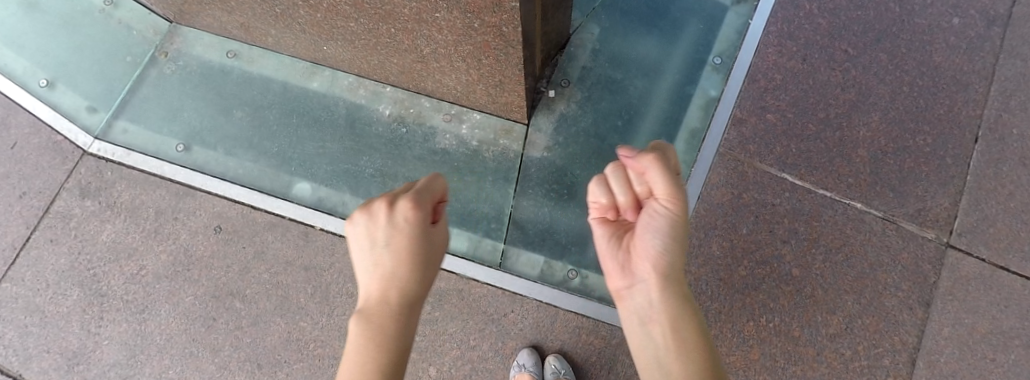}
\includegraphics[width=30mm,height=15mm]{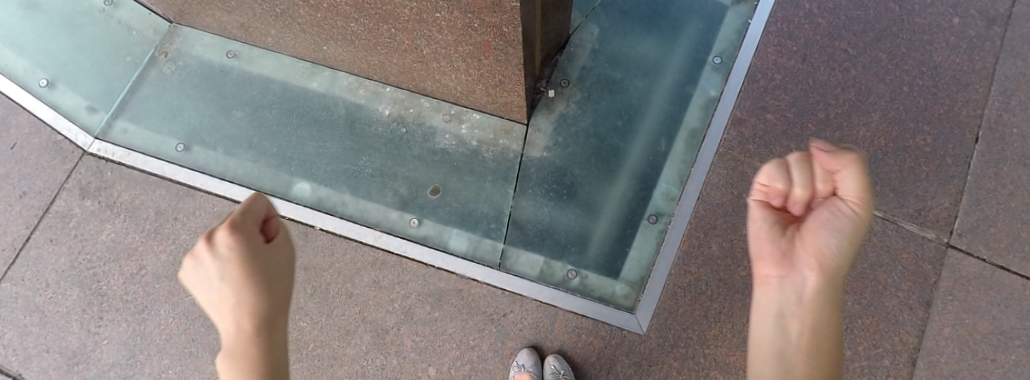}
    \caption{The sequence of our calibration gesture that we bootstrap to obtain ground truth hand segmentation masks to learn a user specific hand appearance network.}
    \label{fig:gesture}
\end{figure*}

\section{Our Approach}

We focus on exploring the benefits of using cooperative human interaction to address the challenges of detecting hands across a diverse set of usage scenarios. In particular, the \textit{training stage} of our method includes two steps to realize gesture-based bootstrapping for pixel-level hand and arm region segmentation. The first step is designed to be able to accurately segment hand regions in any video that demonstrates the predefined gesture. Therefore, we only utilize motion information in the first step to achieve generalization across different users and varied environmental conditions. Specifically, we train gesture-based hand segmentation network by taking optical flow motion field and foreground probability map from background subtraction as input. Since we have strong priors over how the gesture will be performed, the trained detector is robust to any hand shape as well as hand appearance (\eg, the user can even be wearing a glove). In the second step, we utilize appearance network to train the hand region detector with images collected from one gesture interaction video and pixel-level notations from the output of the gesture network. The detector is user-specific as the appearance network only utilizes the information of the exact user in the particular illumination setting. At \textit{test stage}, the hand detector from the second training step is used to detect hands with unconstrained motions. An overview of our proposed interactive hand segmentation framework is illustrated in Figure \ref{fig:network}. 


\subsection{Gesture-based Hand Segmentation} \label{sec:gesture}

We utilize gesture-based hand segmentation to automatically gather "ground truth" segmentation masks of any specific user's hands for training appearance network downstream. In order to achieve generalization across different users and various environment, we only use motion information which is extracted with foreground segmentation methods. While foreground segmentation can segment moving hand regions, the direct results may be corrupted by the noise caused by unavoidable ego-motion of head-mounted cameras, which must be solved as the hand masks from the first stage is served as the proxy 'ground truth' on which the second stage appearance network will be trained. To increase the accuracy, the gesture network, which takes motion clues as input, is used to generate precise hand masks. Moreover, our gesture network quantifies the output's uncertainty which contributes to a significant improvement on the performance of appearance network. 

\textbf{Predefined Gesture:} Utilizing predefined gesture is critical to the success of gesture-based hand segmentation for two reasons: First, it provides relatively stable features so that we can train the gesture network for once and utilize the trained model in any other scenario; Second, the well-designed gesture ensures that the short interaction video includes enough hand information for training appearance network. We consider three aspects while designing the gesture: (1) From the perspective of the user, the gesture should require the absolute minimum amount of effort to perform. (2) From a modeling point of view, the gesture should include both sides of the hand in order to account for the variance in the user's skin tone, as the palm (inner) surface is typically much lighter than the dorsal (outer) surface. (3) From an algorithmic perspective, the gesture should include motions that are easy for motion-based methods to analyze, \eg, not too fast, not too slow and does not cause the body to move too much. Based on these desiderata, the hand gesture is designed to include the following two steps: (1) horizontal hand translation from side to middle with open palms, and (2) hand translation back from the center to side with closed fists. Figure \ref{fig:gesture} illustrates this calibration gesture sequence.


\textbf{Motion Clues:} Since hands are the only moving objects following the path in accordance to the gesture instructions given to the user, we utilize motion-based approach to obtain motion information for segmenting hand regions. We considered two methods to extract motion clues: optical flow and background subtraction. For optical flow, we utilize Bischof's dual TV-L1 optical flow \cite{Zach07aduality} to get dense motion flow fields based on the brightness constancy constraint. Regions of high motion magnitude correspond to hand regions when the hand is moving. We implement the background subtraction method proposed in \cite{bg_sub} which enables statistical estimation of image background by applying Bayesian inference to model a pixel's probability of being a hand.

\textbf{Gesture Network:}Although foreground segmentation methods generally achieve reliable results under ideal conditions, each suffers from its own limitation. To overcome these drawbacks and refine the segmentation masks, we propose a Gesture Network based on \cite{liu2015parsenet} (a semantic segmentation convolutional neural network). The Gesture Network fuses the coarse segmentation results of optical flow field and background subtraction to estimate the hand region segmentation map. This network is trained on manually annotated pixel-level hand masks from a video dataset of calibration gestures performed across many different environments. Since the trained network is not dependent on the environment or appearance of the hand, the training process only needs to be performed once and the trained model can be used for other users in different environment. During the process of use, the gesture network with trained model takes one interactive video as input, and outputs a heat map over all pixels representing the probability of each pixel belonging to hand region. 

\begin{figure}[tb]
    \centering
    \includegraphics[width=75mm,height=15mm]{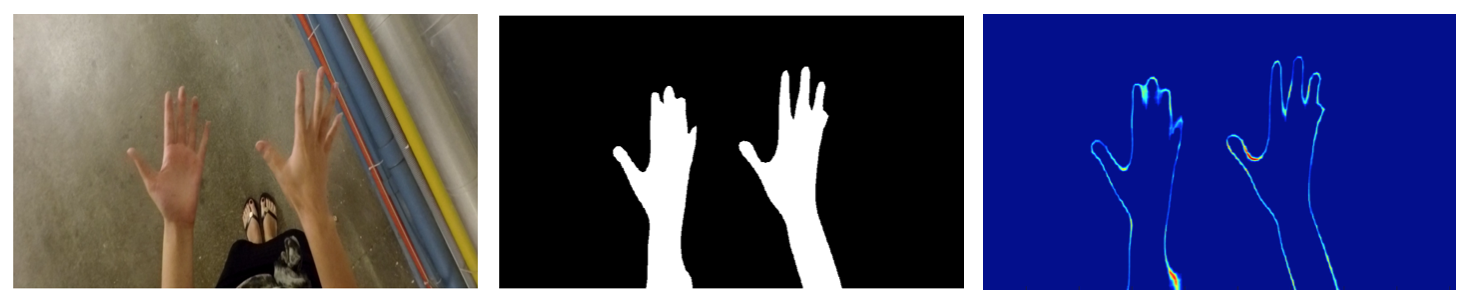}
    \caption{Visualization of output mask and variance over 100 samples.From left to right is original image, expectation mask, and variance mask}
    \label{fig:exp_var}
\end{figure}

\textbf{Characterizing Uncertainty:}\label{sec:uncertainty} Unlike manually labeled masks, the hand masks from gesture network may contain errors that wreck the training of appearance network. Thus, we are interested in estimating the uncertainty in the output of the gesture network. To quantify the confidence/uncertainty at each pixel of the hand detection heat map results, we use dropout \cite{srivastava2014dropout} at test time in the gesture network. Specifically, we apply dropout to various layers of the Gesture Network during training. Instead of calculating the 'ensemble average' of the neuron activations at test time, we also apply dropout at test time running the same input through the gesture network multiple times. By observing the changes in estimate of each pixel over time, we can obtain a Bayesian estimate of the covariance \cite{gal2015dropout}. Figure \ref{fig:exp_var} gives a visual example of the mean and variance of all pixels in a single image. The figure shows high variance in the space between fingers and near the contour of the hand, which is reasonable as those parts are often classified incorrectly.


\subsection{Appearance-based Hand Segmentation}\label{sec:appearance}

With 'ground truth' output of the bootstrapped motion-based gesture network, we are now ready to train an appearance-based hand segmentation network which will be able to detect hands under any type of motion. Recall that we aim to train an appearance-based hand detector that works for a specific user (not all possible users). Accompanied with 'ground truth' pixel notations from output of the gesture network, our appearance network takes the frames in gestural interaction video as input, and learns customized features of one particular user's hands. We leverage the benefit of convolutional neural network and use a modified version of ParseNet\cite{liu2015parsenet} as our appearance network to output a two class probability map.

In order to use uncertainty of the gesture network, we design a weighted loss function using the precision matrix (inverse of the covariance matrix).
$$L = (\sigma(x)-t)^{T}\Sigma^{-1}(\sigma(x)-t)$$
where $L$ is the loss function, $\Sigma^{-1}$ represents the precision matrix, $t$ is pixel notation in the output of gestural network, $x$ is the output of appearance network, and $\sigma(x)$ is sigmoid function $\sigma(x) = \frac{1}{1+e^{-\alpha x}}$.

The partial derivative of loss function with respect to $x$ is
$$\frac{\partial L}{\partial x} = \alpha  \sigma(x)(1-\sigma(x))(\sigma(x)-t)^{T}\Sigma^{-1}$$

We know that the sigmoid function saturates fast. When $\sigma(x)$ is close to 1, the term $(1-\sigma(x))$ will be close to 0, resulting in the magnitude of derivative that is very small. The model is difficult to train under this condition. Therefore, we defined a soft sigmoid function $\sigma(x) = \frac{1}{1+e^{-\alpha x}}$, where the value of $\alpha$ is less than 1. This soft version of the sigmoid function efficiently eliminates saturation, and also helps to speed up the training process.

In general, deep convolutional neural networks require a large amount of data to train the model. In our case, a typical gesture demonstration takes about 6 seconds and we extract about 180 consecutive frames from the center of the temporal window. The temporal window can easily be detected as the user is instructed to remove their hands from the view of the camera before and after the gesture motion. To address the limited amount of training data, we apply two strategies. First, we aggressively regularize the network through the use of early dropout during training. This is in contrast to standard ParseNet training which usually implement dropout layers only after the fully connected layers. By dropping out neurons in the convolutional layers, our appearance network is able to achieve optimal performance. Second, we implement data augmentation to prevent over-fitting. Since the background during the gesture demonstration is limited to one scene, we also explore background augmentation by including other images of the surrounding environment.

\section{Experimental Results}

\subsection{Personalized Hand Detection Dataset}

Since existing egocentric activity datasets \cite{fathi2011learning} \cite{li2013pixel} do not contain any consistent gestural inputs, we created a new dataset that is more appropriate for our interactive scenario. The dataset includes 10 users in 30 different environmental conditions (Figure \ref{fig:dataset}), which is designed to cover skin tones from light to dark, environment setting from indoor to outdoor. Furthermore, the dataset also includes hand object interaction under extreme conditions, \eg hands in very dim light, tasks with bandaged hands. For each user in each scenario, there are videos for training with gestural demonstrations, and corresponding test videos including a wide range of various body and hand motions, all of which are accompanied with pixel-level manually labeled hand masks for comparing results. To train our appearance-based baseline models, we utilize two publicly available benchmark datasets, GTEA \cite{fathi2011learning} and CMU-EDSH \cite{li2013pixel} as the images are taken from first-person point-of-view and include interactions between hands and objects in varied forms.


\subsection{Gesture Network Evaluation}

\begin{table}[tb]
\begin{center}
\begin{tabular}{|l|c|}
\hline
Input combination & F1 score\\
\hline\hline
Bg-sub & 0.924 \\
Opt & 0.919 \\
Bg-sub + Opt x-direction & 0.928\\
Bg-sub + Opt y-direction & 0.931\\
Bg-sub + Opt x-direction + Opt y-direction & \textbf{0.946}\\
\hline
\end{tabular}
\end{center}
\caption{Hand segmentation performance with different input combination, where Bg-sub represents foreground probability map calculated from background subtraction, and Opt represents optical flow}
\label{table:gesture_input}
\end{table}

\subsubsection{Ablative Analysis}

As discussed in Section \ref{sec:gesture}, the video input of the gesture network is designed to capture the motion information of a predetermined gesture. In this experiment, we seek to understand which motion cues deliver most information for training the gesture network and show best performance while detecting hand regions during a gesture demonstration. Therefore, we perform ablative analysis across three input types: (1) background subtraction, (2) dense optical flow in the x direction and (3) dense optical flow in the y direction. We design five experiments using different combination of three motion-based input types. Table \ref{table:gesture_input} contains the F1 scores with respect to different combination of input types. The results show that best hand detection performance are obtained while using background subtraction and optical flow in both directions.

\begin{figure}[tb]
    \centering
    \includegraphics[width=25mm,height=15mm]{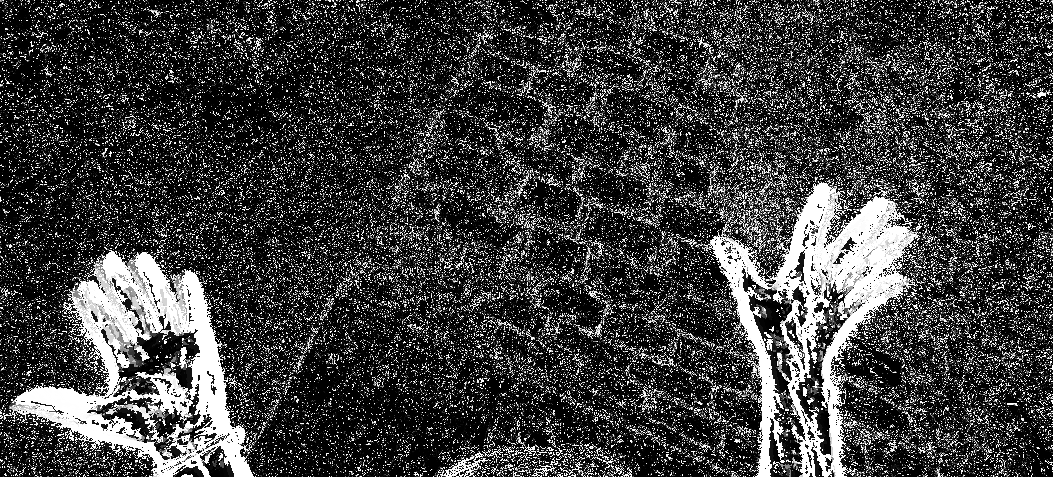}
    \includegraphics[width=25mm,height=15mm]{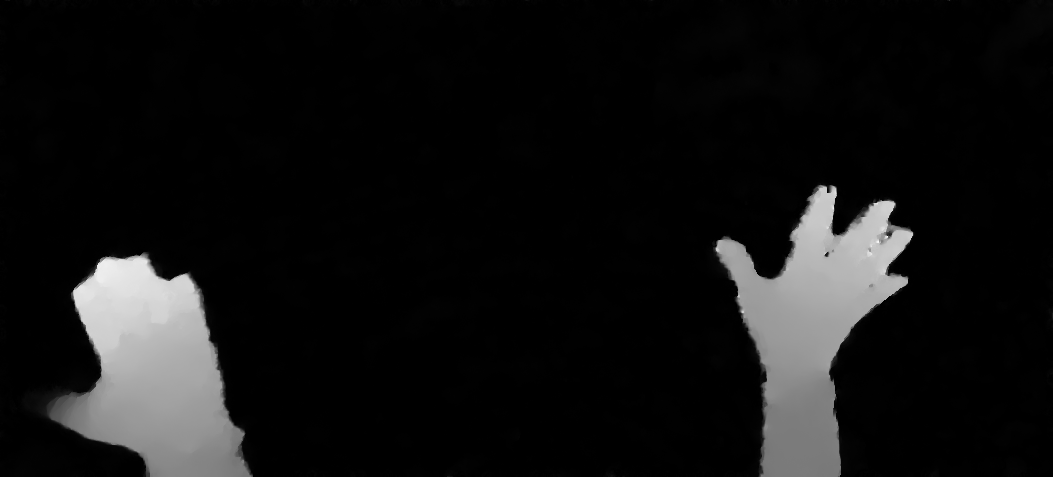}
    \includegraphics[width=25mm,height=15mm]{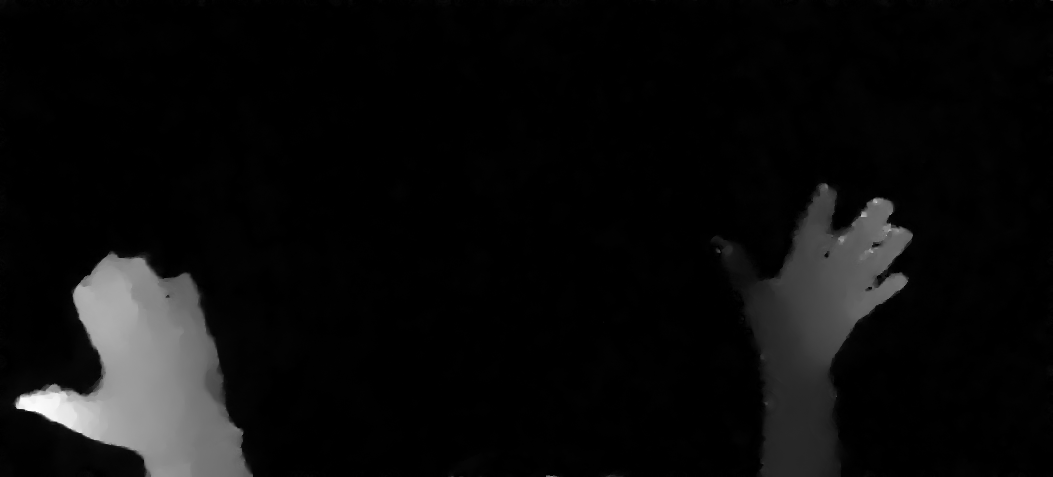}
    \caption{Visualization of input for gesture network}
    \label{fig:gesture_input}
\end{figure}

Figure \ref{fig:gesture_input} shows sample images of the foreground probability map computed from background subtraction and the optical flow field in the $x$ and $y$ direction, respectively. As one would expect, the optical flow works well on segmenting hand regions when the hands are in motion and the background subtraction is able to find hand contours even when there is very little motion.

\subsubsection{Implementation Details} 

\textbf{Motion Information:} As discussed in paper, the combination of background subtraction and optical flow along both directions as input yields best performance for gesture-based hand detection. To generate the background subtraction input, we utilize the method in \cite{Godbehere2012Visualtrack} with one initialization frame. The prior probability of each pixel being background is $p=0.8$ according to our rough estimate of the ratio of background to hand pixels in gesture interaction videos, and the smoothing ratio and learning rate are 0 and 0.6 respectively. For the optical flow component of the input, we utilize dual TV-L1 optical flow \cite{Zach07aduality} with smoothness parameter $\lambda=0.15$, stop criterion parameter $\epsilon=0.01$, and iteration numbers up to 300 for optimal performance.

\textbf{Network Implementation:} Our gesture network architecture is based on ParseNet \cite{liu2015parsenet} which contains five convolution layers and two fully connected layers. We introduce dropout after the convolutional layers from conv3 to conv5 with a dropout ratio of 0.4 both during training and testing phases. The loss function is per-pixel weighted two-class softmax loss with hand and non-hand pixels weighted by 5 and 0.6, respectively. During training step, the input size is $K \times C \times 380 \times 1030$, where $K$ represents batch size and always equals to 1, $C$ equals to 3, the number of channels corresponding to statistical estimate of heat map from background subtraction and optical flow in two dimensions. The learning rate is initialized to $10^{-8}$ and we use polynomial decaying learning rate schedule to train our network. During test time, we pass each input through the gesture network 100 times to obtain mean and covariance estimate for the output of gesture network.

\subsection{Appearance Network Evaluation}

\subsubsection{Network Implementation}

The architecture of appearance network is based on the protocol of ParseNet \cite{liu2015parsenet} which contains five convolution layers and two fully connected layers. We introduce dropout after convolution layers from conv3 to conv5 along with dropout between the fully connected layers. As described in the paper the loss function is the per-pixel weighted Euclidean loss accounting for the uncertainty in the gesture network. The appearance network takes 3 seconds interval (around 180 frames) of the gesture interaction video as input, and trains the personalized hand detector. The training policy is polynomial decaying with learning rate initialized to $10^{-8}$. The input size is $K \times C \times 380 \times 1030$, where $K$ represents batch size and always equals to 1. $C$ equals to 3 for R, G, B channels in one frame. 





\subsubsection{Data Augmentation Analysis}

Based on the features of our training set, we apply three data augmentation strategies. We implement \emph{Transformation augmentation} with cropping, rotation and flipping to generate samples with more spatial variety. First, we randomly crop half of the images to size $304 \times 824$, then scale back to $380 \times 1030$. To enhance model ability of learning hands in different direction, we randomly apply rotation transformation with angle $\pm 30^\circ,\pm 15^\circ$. We also randomly flip the images horizontally.

\begin{figure}[tb]
\small
\center
\includegraphics[width=35mm,height=17mm]{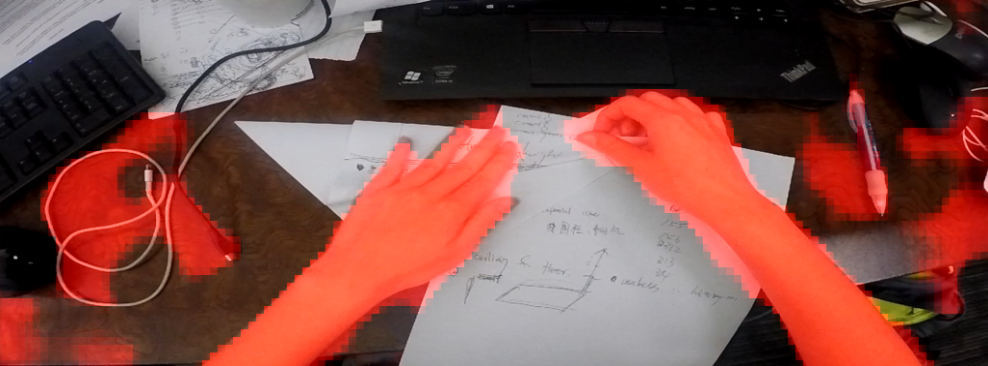}
\includegraphics[width=35mm,height=17mm]{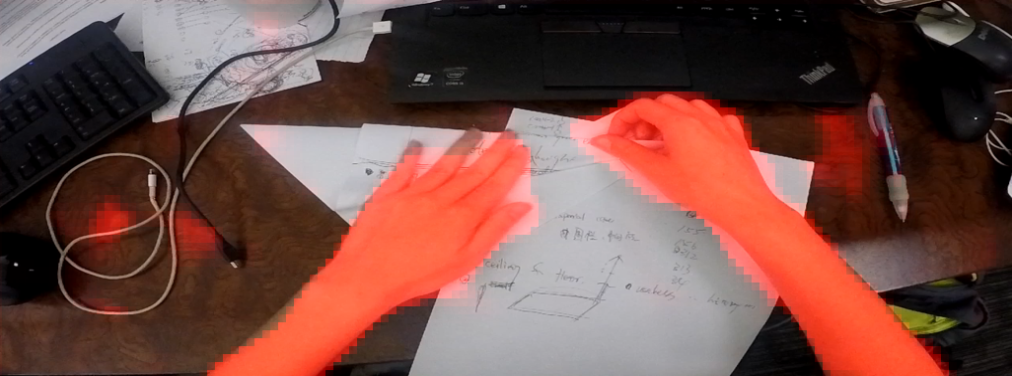}
\\ (a) before augmentation \hspace{8mm} (b) after augmentation
\caption{Background Augmentation Results.}
\label{fig:bg_aug}
\end{figure}

Due to lack of background diversity in gestural interaction videos, we implement \emph{background augmentation} by including 30 images without hands. We collect images by instructing the user to look around before performing gesture. In order to give intuitive grasp of the performance of background augmentation, we visualize two output images from models trained with and without background images in Figure \ref{fig:bg_aug}. 

In order to enhance the model to adapt to illumination variation, we apply \emph{brightness augmentation} on all images. We first calibrate the brightness level by transforming the images from the RGB color model to HSV color space where the value (V) represents brightness. We scale "V" term to be 5 levels from 0.2 to 0.6, which simulates images in different illumination conditions. We analyze the performance of different data augmentation strategies on one test set with 177 images in Table\ref{table:aug}. 

\begin{table}[tb]
\begin{center}
\small
\begin{tabular}{|l|c|}
\hline
Data Augmentation strategy & F1 score\\
\hline\hline
No augmentation& 0.834\\
Brightness & 0.852 \\
Environment & 0.867\\
Brightness + Transformation & 0.854 \\
Environment + Transformation & 0.865 \\
Brightness + Transformation + Environment & \textbf{0.869}\\
\hline
\end{tabular}
\end{center}
\caption{Performance evaluation of augmentation strategy}
\label{table:aug}
\end{table}

\subsubsection{Impact of Dropout on Training}

As described in \cite{srivastava2014dropout}, dropout is a powerful technique to avoid over-fitting. By randomly removing units with certain ratio during training, the network achieves better performance as well as robustness to small amount of data. In our case, the number of input images is around 180. Therefore, we implement dropout layers at early stages to tackle the problem of limited training samples. We show the result on one test set with 177 images in Table \ref{table:dropout}.

\begin{table}[]
\centering
\footnotesize
\begin{tabular}{|l|c|c|}
\hline
\multirow{2}{*}{Layers with Dropout}              & \multicolumn{2}{l|}{Dropout ratio} \\ \cline{2-3} 
                                                  & 0.4              & 0.5             \\ \hline\hline
fc7 + fc6                                         & 0.854            & 0.853           \\ \hline
fc7 + fc6 + conv5                                 & 0.853            & 0.854           \\ \hline
fc7 + fc6 + conv5 + conv4                         & 0.863            & 0.860           \\ \hline
fc7 + fc6 + conv5 + conv4 + conv3                 & 0.855            & 0.856           \\ \hline
fc7 + fc6 + conv5 + conv4 + conv3 + conv2         & 0.842            & 0.843           \\ \hline
fc7 + fc6 + conv5 + conv4 + conv3 + conv2 + conv1 & 0.799            & 0.782           \\ \hline
\end{tabular}
\caption{Performance comparison ($F_1$ score) using dropout.}
\label{table:dropout}
\end{table}

\begin{figure*}[t]
\centering
\includegraphics[width=30mm,height=12mm]{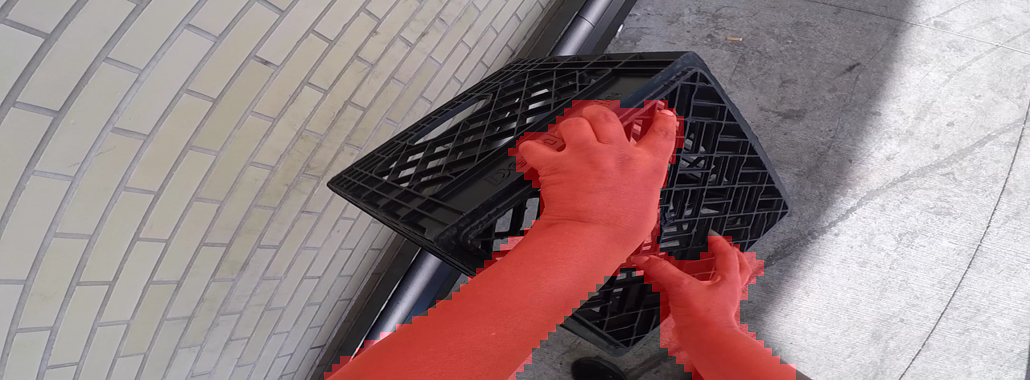}
\includegraphics[width=30mm,height=12mm]{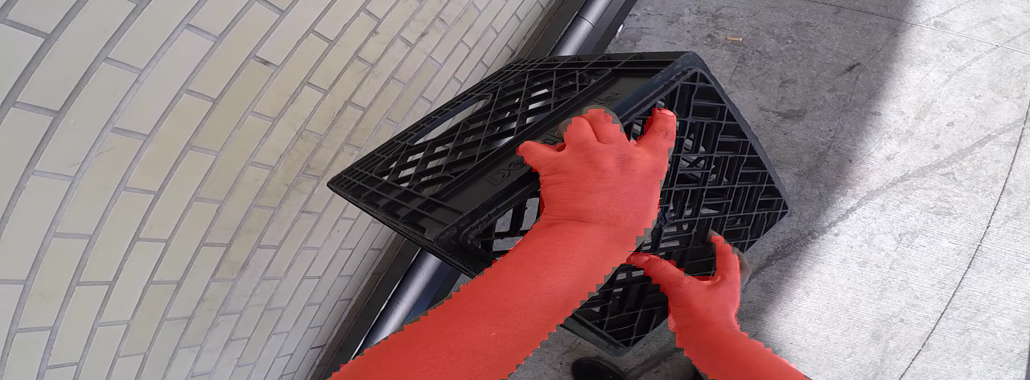}
\includegraphics[width=30mm,height=12mm]{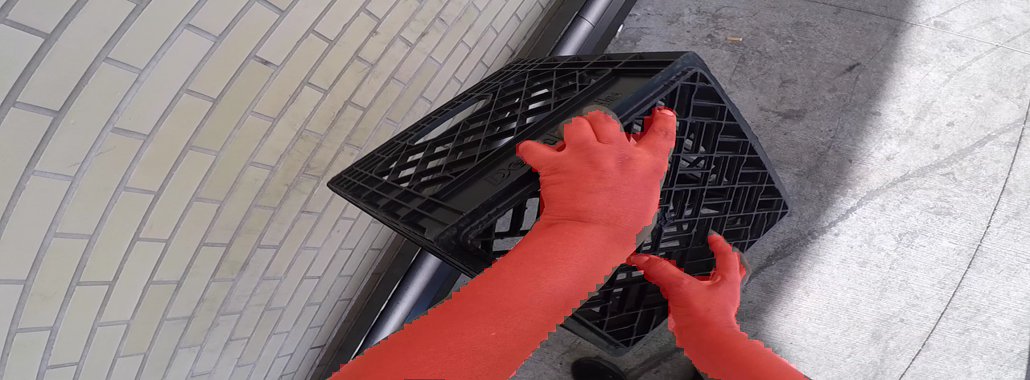}
\includegraphics[width=30mm,height=12mm]{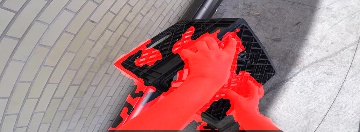}
\includegraphics[width=30mm,height=12mm]{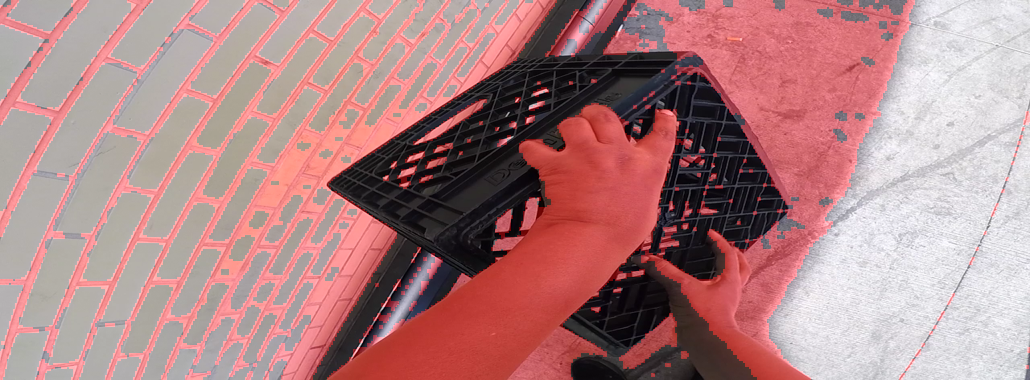}
\\
    \includegraphics[width=30mm,height=12mm]{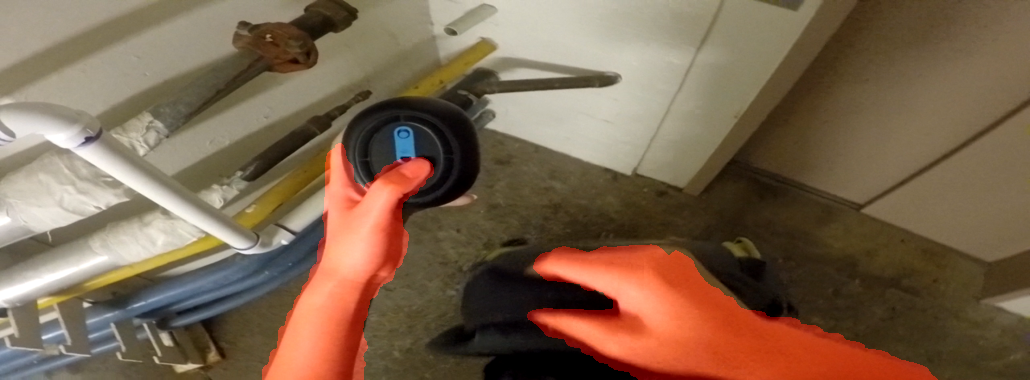}
    \includegraphics[width=30mm,height=12mm]{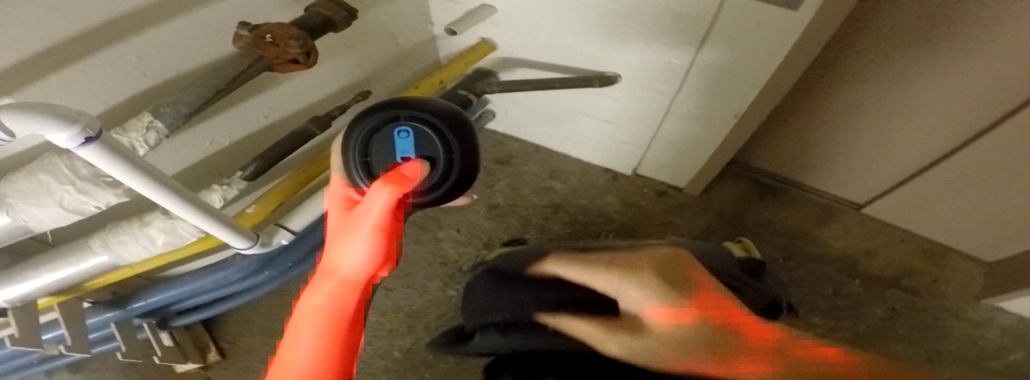}
    \includegraphics[width=30mm,height=12mm]{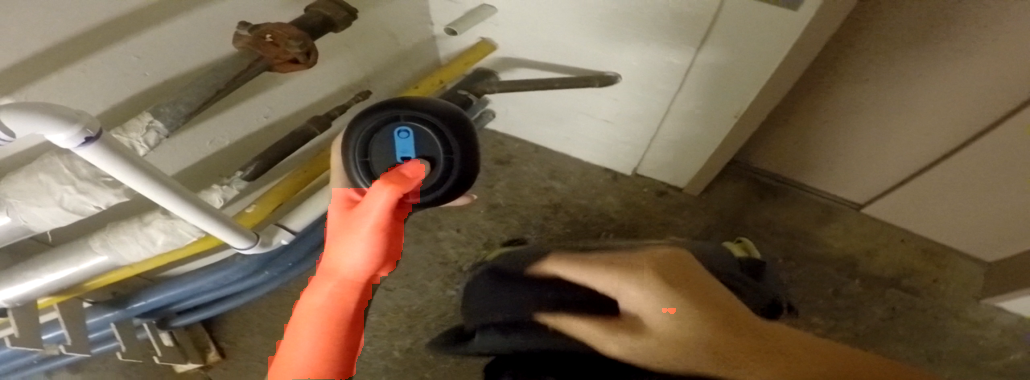}
    \includegraphics[width=30mm,height=12mm]{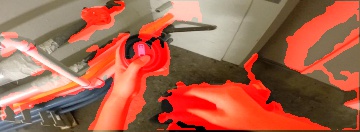}
    \includegraphics[width=30mm,height=12mm]{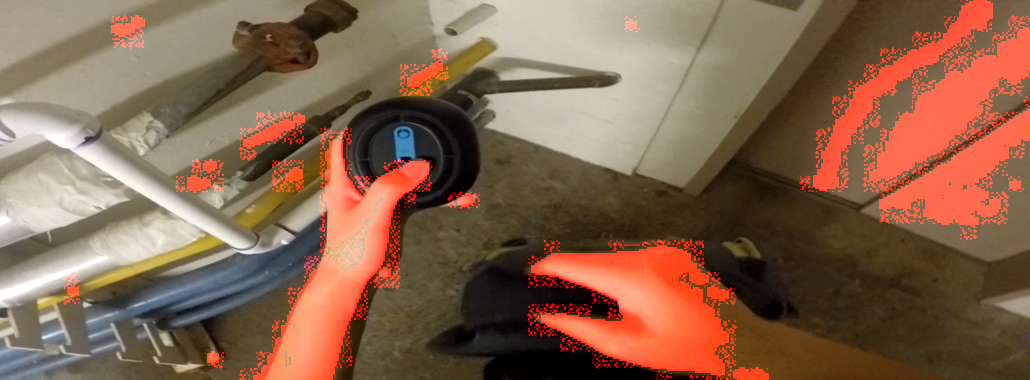}
\\
    \includegraphics[width=30mm,height=12mm]{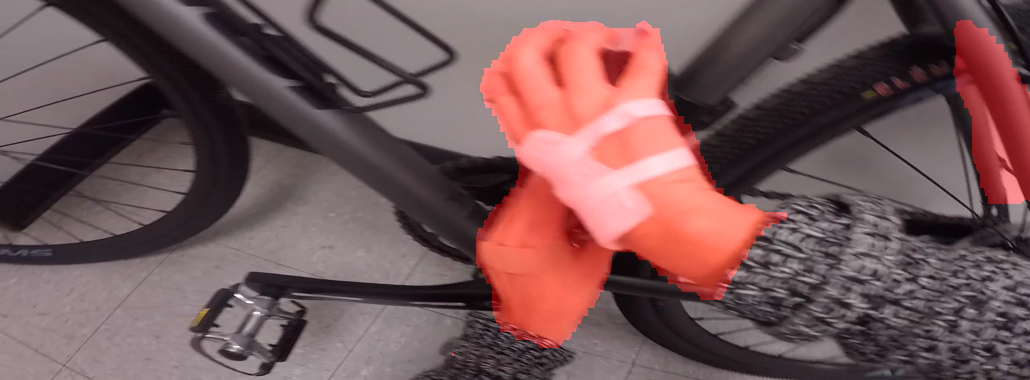}
    \includegraphics[width=30mm,height=12mm]{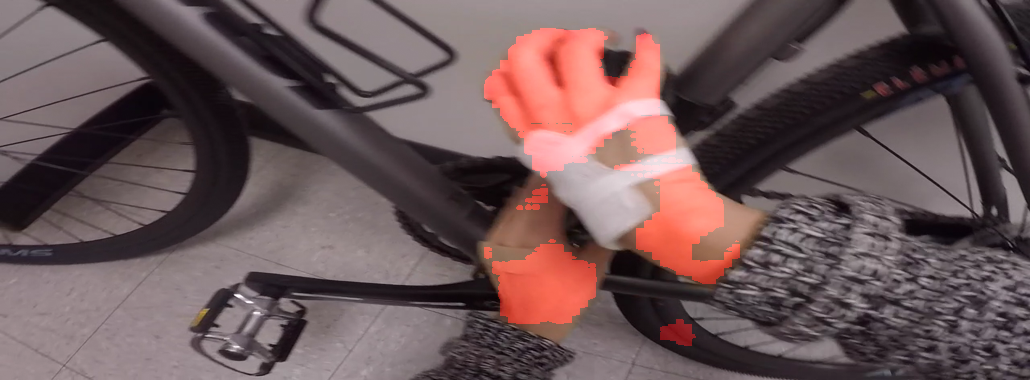}
    \includegraphics[width=30mm,height=12mm]{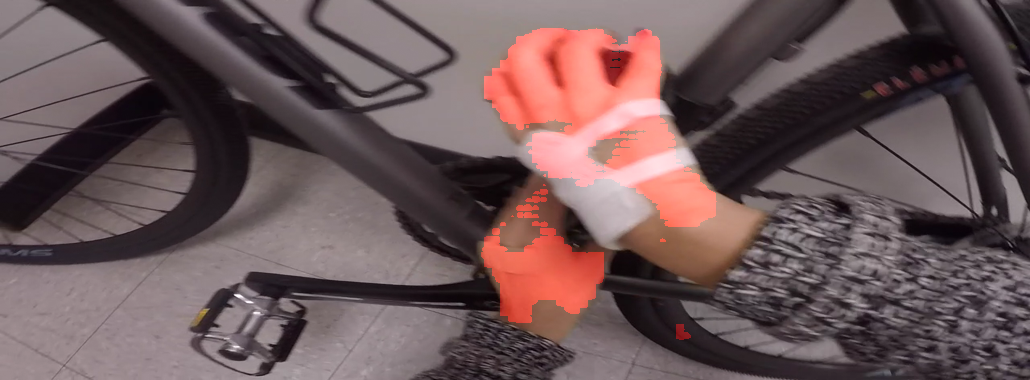}
    \includegraphics[width=30mm,height=12mm]{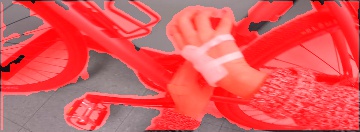}
    \includegraphics[width=30mm,height=12mm]{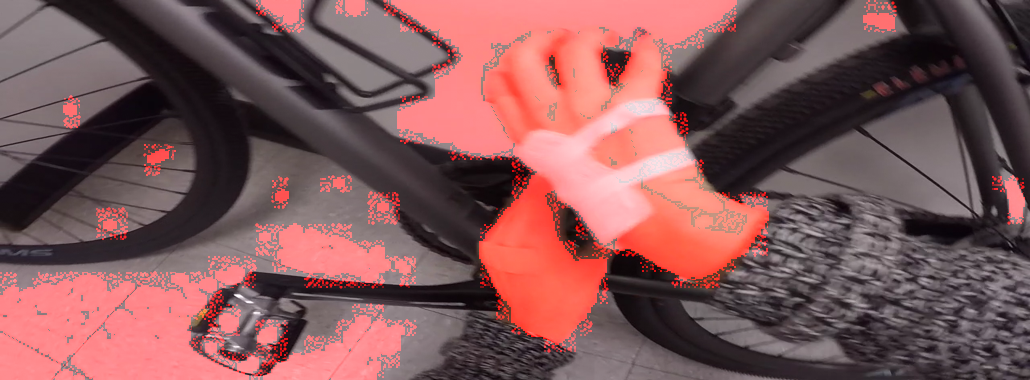}
\\
    \includegraphics[width=30mm,height=12mm]{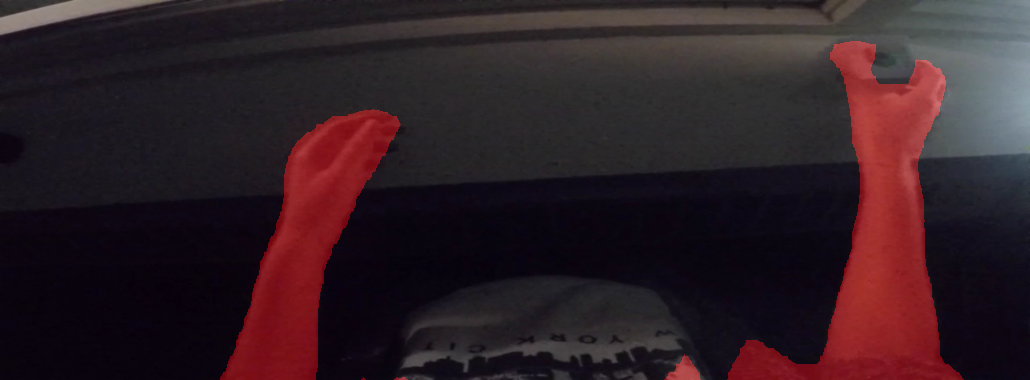}
    \includegraphics[width=30mm,height=12mm]{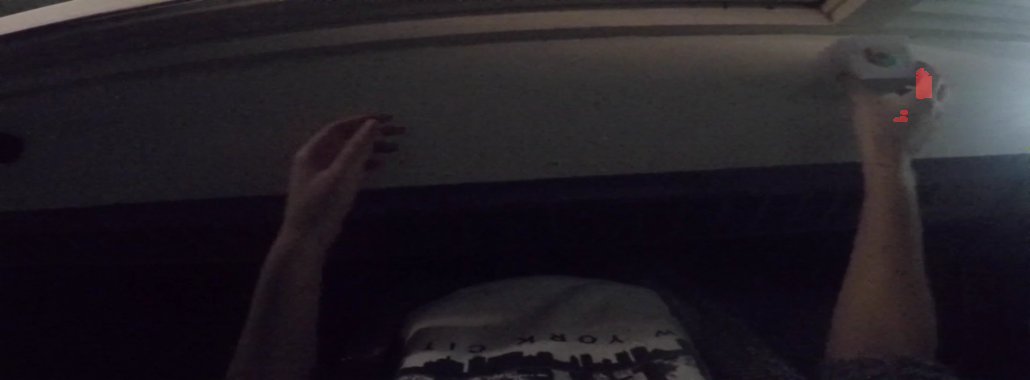}
    \includegraphics[width=30mm,height=12mm]{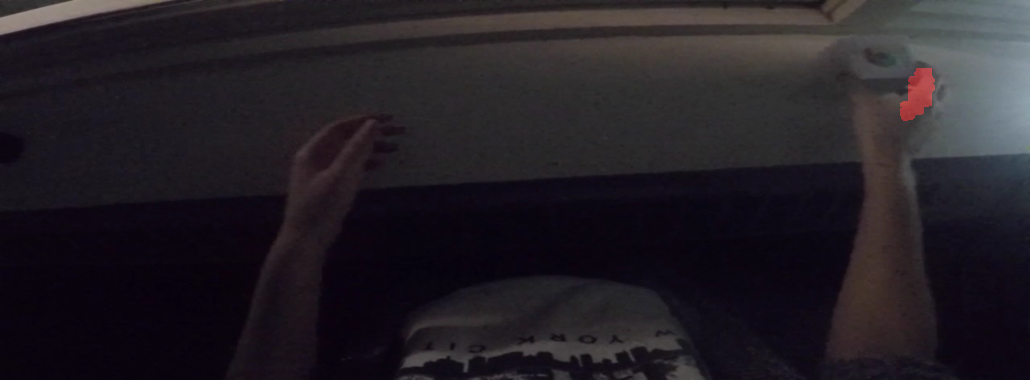}
    \includegraphics[width=30mm,height=12mm]{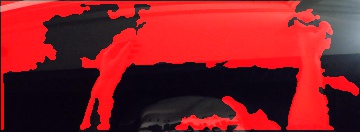}
    \includegraphics[width=30mm,height=12mm]{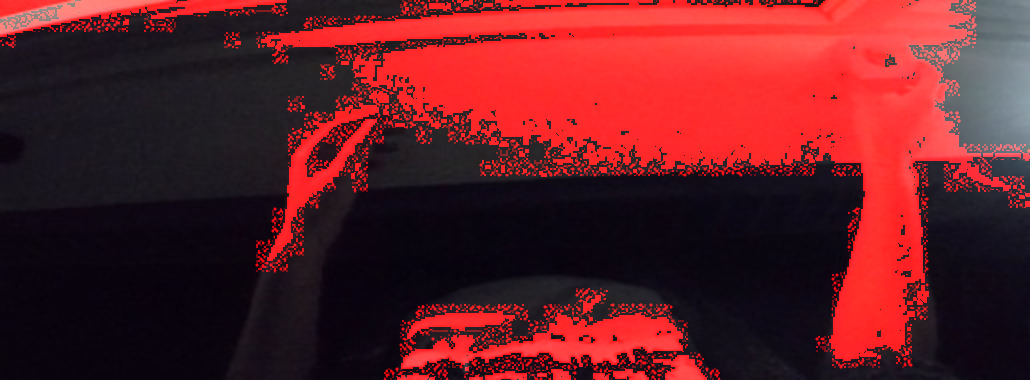}
\\
    \includegraphics[width=30mm,height=12mm]{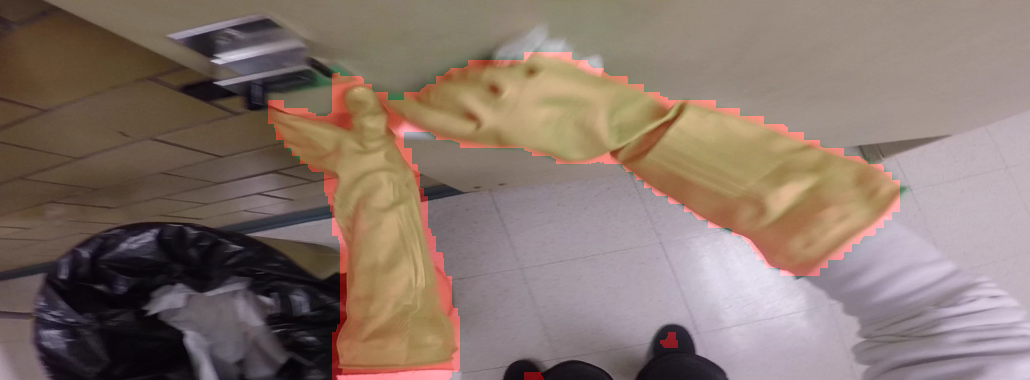}
    \includegraphics[width=30mm,height=12mm]{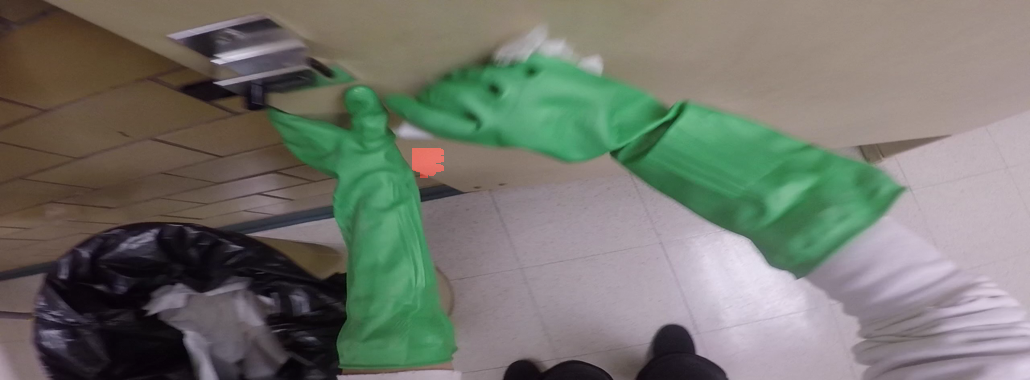}
    \includegraphics[width=30mm,height=12mm]{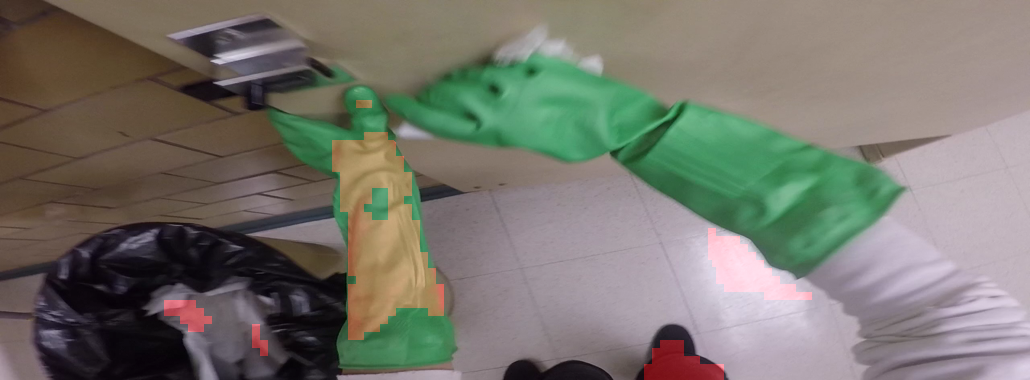}
    \includegraphics[width=30mm,height=12mm]{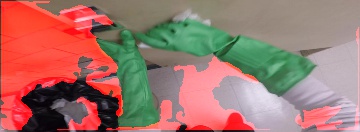}
    \includegraphics[width=30mm,height=12mm]{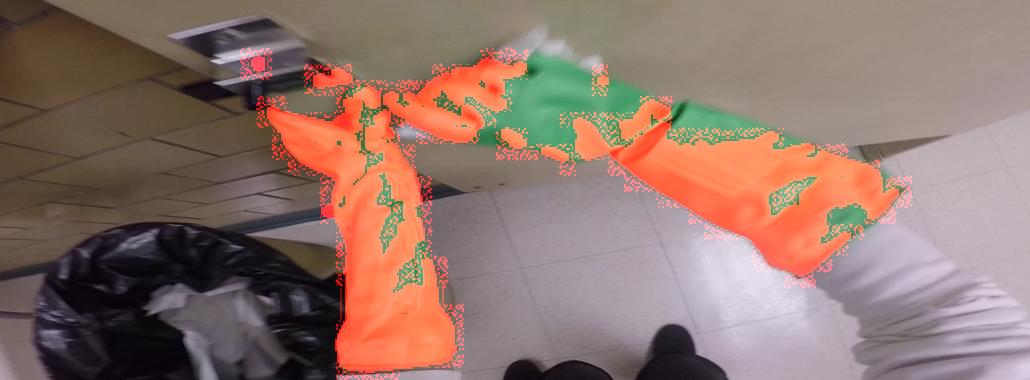}
\\
\centering
(a) Proposed \hspace{1.4cm}
(b) ParseNet \hspace{1.4cm}
(c) FCN \hspace{1.4cm}
(d) PerPix \hspace{1.4cm}
(e) GMM\cite{kumar2015fly}
\caption{Output visualization of proposed method and baselines}
\label{fig:baseline_comparison}
\end{figure*}

\subsubsection{Data Dependency Analysis} 

Given the fact that one gesture interaction video only contains around 180 images, we seek to explore the benefit of including multiple gesture interaction videos while training the appearance network since more videos will potentially increase data diversity to benefit the model. We show the performance of increasing gestural interaction video numbers in Table \ref{table:video_num}. The evaluation result shows that in the certain range of video numbers, the accuracy of the model increases with more calibration videos for training. 

\begin{table}[tb]
\footnotesize
\begin{center}
\begin{tabular}{|l|c|c|c|c|c|c|c|}
\hline
Vid. & 1&2&3&4&5&6&7\\
\hline\hline
F1 & 0.823& 0.844& 0.842& 0.868& 0.860& 0.876& 0.875 \\
\hline
\end{tabular}
\end{center}
\caption{Performance of multiple gestural interaction videos as input}
\label{table:video_num}
\end{table}

\subsubsection{Impact of Modeling Uncertainty}

We defined the loss function of the appearance network in Section \ref{sec:appearance} which takes account of the precision matrix generated from gesture network. The precision matrix is diagonal, which is passed into the appearance network as input. In order to evaluate the effect of considering uncertainty, we train two appearance networks with loss function that utilizes identity matrix or precision matrix. The F1 scores of the outputs that consider uncertainty or not are 0.842 and 0.859 respectively, which proves that by introducing precision term in loss function, the performance of appearance network improves.


\begin{table}
\begin{center}
\begin{tabular}{|l|c|c|c|c|c|}
\hline
Method & test1 & test2 & test3 & test4 & test5\\
\hline\hline
\textbf{Proposed} & 0.873 & \textbf{0.853} & \textbf{0.803} & \textbf{0.842} & \textbf{0.897}  \\
ParseNet \cite{liu2015parsenet} & \textbf{0.886} & 0.852 & 0.598 & 0.203 &0.01 \\
FCN \cite{long2015fully}& 0.872 & 0.845 & 0.572 & 0.213 &0.172 \\
PerPix \cite{li2013pixel}& 0.453 & 0.661 & 0.625 & 0.003 &0.002\\
GMM \cite{kumar2015fly} &0.439 & 0.736 &0.213 &0.326&0.732\\
\hline
\end{tabular}
\end{center}
\caption{F1 score of hand segmentation performance against baseline methods}
\label{table:baseline_comparison}
\end{table}

\subsection{Test stage Evaluation}

We present the performance comparison of our proposed method against several baselines in Table \ref{table:baseline_comparison}. We report results for five test sets, each including 50 images. The test 1 to test 5 corresponds to the test sets of outdoor environment, indoor environment, dark environment, hand with bandage and tasks with gloves. The first two are under normal conditions while the last three represent extreme conditions which are hard tasks for hand detection. According to the quantitative comparison, a standard convolutional neural network achieves remarkable performance on hand segmentation under conditions where normal hand features are easy to track. However, under extreme conditions \eg work in very dark room or action with gloves, the pretrained models will fail due to a large difference between the training distribution and the test distribution. In contrast, our personalized detector performs well with the help of gestural bootstrapping. The visualization of the output is illustrated in Figure \ref{fig:baseline_comparison}.

As a further comparison of our method and baselines, we point out the two salient features of our framework: (1) the idea of performing gesture-based bootstrapping to enhance generalization of hand segmentation tasks on detection step, and (2) introducing a convolutional neural network as an efficient tool for pixel level hand detection. The GMM-based method \cite{kumar2015fly} also introduces human cooperative interaction to train different detectors according to varied scenarios. However, it does not explicitly separate the motion and appearance processes. As a result, it is vulnerable to scenarios where the skin appearance is not strictly included inside the color spaces that it pre-defines, and where the background in test scenario is similar to hand appearance. Also, since the detector is only trained with features in typical hand region, it does not include some necessary information, \eg shadows on hands, background environments. Moreover, GMM-based method needs to tune parameters for each different scenario to get optimal results, which is not practical for real-time implementation. Our method, by fully leveraging the benefits of human interaction and through the modeling capacity of convolutional neural networks, proposes an appearance-independent gesture interaction step that is both stable and efficient across a wide variety of hand and background appearances.

Our approach shares some features with the Parsenet \cite{liu2015parsenet} and FCN \cite{long2015fully} baselines in that we also utilize a deep convolutional neural network for hand segmentation tasks in egocentric videos. Convolutional neural networks outperform many other generative models, \eg, pixel level hand detector \cite{li2013pixel}, since it enables learning very complex models for different scenarios, \eg, diverse hand texture appearance across human races, and both indoor and outdoor environments. The fail cases for CNN approaches are in scenarios with extreme conditions, as shown in the paper, \eg, in very dark room, wearing gloves, and with bandage on hands. Instead of letting the training set limit the capacity of the model and necessitate human energy for labeling, our method takes advantage of both cooperative human interaction and deep convolutional networks to provide high performance on a huge variety of datasets.


\section{Conclusion}

In this paper, we proposed  a method that uses human interaction to bootstrap the hand segmentation algorithm. The method includes two stages, where the first step utilizes human interaction to realize gesture-based bootstrapping to generate target labels for training  a discriminative appearance-based hand detector in the second step. We performed several experiments in conditions across a wide range of illumination and hand appearance, and showed that our model is both robust and more accurate even in the most challenging environments.

{\small
\bibliographystyle{ieee}
\bibliography{egbib}
}

\end{document}